\documentclass{article}

\usepackage[final]{neurips_2024}

\usepackage[utf8]{inputenc} 
\usepackage[T1]{fontenc}    
\usepackage{hyperref}       
\usepackage{url}            
\usepackage{booktabs}       
\usepackage{amsfonts}       
\usepackage{nicefrac}       
\usepackage{microtype}      
\usepackage{enumitem}

\usepackage{multirow}
\usepackage{amsmath}
\usepackage{subcaption}
\usepackage{tabularx}
\usepackage{wrapfig}
\usepackage{float}

\usepackage{xcolor}         
\definecolor{darkgreen}{rgb}{0.0, 0.4, 0.0}

\usepackage[colorinlistoftodos,prependcaption,textsize=tiny]{todonotes}
\usepackage{xargs}
\hypersetup{
    colorlinks=true,
    linkcolor=blue,       
    citecolor=blue,      
    urlcolor=cyan,        
    filecolor=magenta,    
    linktocpage=true,
}

\usepackage{makecell}

\title{
Comparing Bottom-Up and Top-Down Steering Approaches on In-Context Learning Tasks
}

\author{%
    Madeline Brumley\textsuperscript{†} \\
    University of Washington \\
    \And
    Joe Kwon \\
    Massachusetts Institute of Technology \\
    \AND
    David Krueger \\
    MILA \& Université de Montréal \\
    \\
    \And
    Dmitrii Krasheninnikov \\
    University of Cambridge \\
    \And
    Usman Anwar\textsuperscript{†} \\
    University of Cambridge
}

\newcommand\thanksline{%
   Correspondance at: \texttt{madbrum@uw.edu, ua237@cam.ac.uk}
}

\makeatletter
\let\ognotice\@noticestring
\g@addto@macro\@noticestring{\par\smallskip\thanksline}
\makeatother

\begin{document}

\maketitle

\begin{abstract}
A key objective of interpretability research on large language models (LLMs) is to develop methods for robustly steering models toward desired behaviors. To this end, two distinct approaches to interpretability -- ``bottom-up" and ``top-down" -- have been presented, but there has been little quantitative comparison between them. We present a case study comparing the effectiveness of representative vector steering methods from each branch: function vectors (FVs)~\citep{todd2024function}, as a bottom-up method, and in-context vectors (ICVs) \citep{liu2024incontext} as a top-down method. While both aim to capture compact representations of broad in-context learning tasks, we find they are effective only on specific types of tasks: ICVs outperform FVs in behavioral shifting, whereas FVs excel in tasks requiring more precision. We discuss the implications for future evaluations of steering methods and for further research into top-down and bottom-up steering given these findings.
\end{abstract}

\section{Introduction}

\begin{wrapfigure}{r}{0.4\textwidth}
    \centering
    \includegraphics[width=0.99\linewidth]{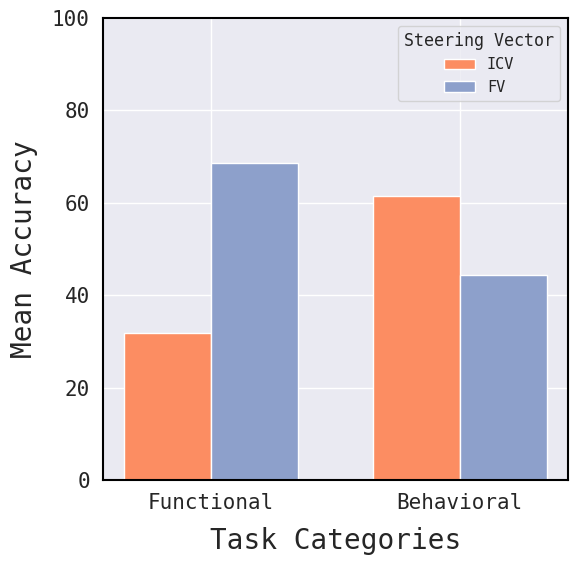}
    \caption{Average performance of in-context vectors \citep{liu2024incontext} and function vectors \citep{todd2024function} on different classes of in-context learning tasks.} 
    \label{fig:summary}
    \vspace{-4em}
\end{wrapfigure}

Vector steering has emerged as a promising method for controlling the behavior of large language models (LLMs), effective at steering toward desirable behaviors such as honesty or away from undesirable behaviors such as sycophancy \citep{zou2023representation, panickssery2024steeringllama2contrastive}.
Vector steering involves creating a \textit{concept vector} that encodes desired behaviors and applying it to the model during inference.
This vector is typically derived from interpretability analyses, which can be broadly categorized into ``bottom-up" and ``top-down" approaches.

Bottom-up interpretability focuses on understanding fine-grained, low-level mechanisms within neural networks, such as individual neurons or circuits \citep{olah2020zoom}, and how these components contribute to the model's overall functionality~\citep{gandelsman2024interpreting}.
In contrast, top-down interpretability examines broader neural representations by analyzing the global activity of neuron populations and the high-level concepts they may represent \citep[][]{zou2023representation}.

Both bottom-up and top-down frameworks for extracting steering vectors show promising results on their respective set of evaluation tasks.
However, it is difficult to meaningfully analyze the general comparative effectiveness of the two classes of methods with existing individual evaluations, since they are evaluated on highly disparate tasks.
Thus, to help understand the relative merits of top-down and bottom-up interpretability, we present a comparative study of two representative vector steering methods -- \emph{in-context vectors} \citep[][]{liu2024incontext} and \emph{function vectors} \citep[][]{todd2024function} -- on a unified set of diverse in-context learning (ICL) tasks. 

In-context vectors are extracted via top-down approach by analyzing differences in activations induced by presenting the model with contrastive examples of a target behavior. 
These vectors capture broader concepts and do not attempt to pinpoint exact neural elements responsible for the behavior of interest.
Function vectors are computed by identifying key attention heads causally responsible for high performance on in-context learning tasks. 
As such, function vectors use a bottom-up approach by targeting the analysis on ``microscopic" neural components responsible for encoding relevant behaviors.

The main findings of our study are as follows:
\begin{itemize}
\item FVs excel at steering precise, fine-grained behaviors but struggle with high-level concepts.
\item Conversely, ICVs are effective at inducing broader behavioral shifts but are less reliable for precise tasks.
\item Additionally, ICVs are more prone to causing undesired degradation in model fluency and lack robustness when applied in different contexts. 
\end{itemize}

Overall, we found that both methods have more considerable limitations than suggested in prior work.
This emphasizes the need for a unified evaluation benchmark for intervention steering methods and suggests that both interpretability approaches currently have important shortcomings.
Our results also hint at the fact that in-context learning behaviors are likely driven by driven mechanisms in different contexts.

\section{Setup}

\label{sec:extract_methods}
\textbf{Computing steering vectors.} To obtain the vectors, we closely follow the procedures outlined by
\citet{todd2024function} for FVs and \citet{liu2024incontext} for ICVs. 
We sweep across different vector strengths and demonstration set sizes to obtain the best ICV for each task. However, we do \textit{not} perform the same sweeps for FVs, as the strength of FVs cannot be controlled by a hyperparameter in the way that ICVs can (see Appendix \ref{sec:eval_details} for further explanation). FVs are added to a single layer $l \approx L/3$ where $L$ is the total number of layers in the model; ICVs are added at all layers of the model. These procedures were found to yield the best results in the respective original studies. See Appendix~\ref{sec:experiment} for more details.

\textbf{Models.} We conduct our main experiments on Llama 2-Chat (7B) \citep{touvron2023llama2openfoundation}, following prior work investigating steering vectors, including \citet{todd2024function}. We also run our evaluations of the detoxification task on pretrained Llama 2 (7B). We use HuggingFace implementations of these models \citep{wolf2020huggingface}. 

\textbf{Tasks.} We evaluate the steering capabilities of both methods across 7 diverse in-context learning tasks, categorized into two groups: (1) functional tasks, which require precise input-output transformations, and (2) behavioral tasks, which test for broader shifts in model behavior or writing style. These tasks are sampled from \citet{todd2024function} and \citet{liu2024incontext}, though some implementations diverge slightly (see Appendix \ref{sec:eval_details}). Datasets for these tasks are discussed in Appendix \ref{sec:datasets}.

The functional tasks are:
\vspace{-1mm}
\begin{itemize}[noitemsep, topsep=0pt]
    \item \textbf{Antonym:} Given an input word, generate a word with the opposite meaning.
    \item \textbf{Capitalize:} Given an input word, capitalize the first letter of the word.
    \item \textbf{Country-Capital:} Given a country, generate its capital city.
    \item \textbf{Synonym:} Given an input word, generate a synonym of the word.
\end{itemize}

The behavioral tasks are:
\vspace{-1mm}
\begin{itemize}[noitemsep, topsep=0pt]
    \item \textbf{Detoxification:} Detoxify a toxic sentence.
    \item \textbf{Sentiment Transfer:} Convert a sentence with negative sentiment into a continuation with positive sentiment.
\end{itemize}

To study the effects and generalization of both methods, we follow \citeauthor{todd2024function} and evaluate functional tasks over zero-shot and shuffled-label 3-shot settings, as well as in out-of-distribution (OOD) natural text settings (see Appendix \ref{sec:experiment}). We evaluate behavioral tasks only on zero-shot settings.

\begin{figure}
    \vspace{-10mm}
    \centering
    \begin{subfigure}[b]{0.48\textwidth}
        \centering
        \includegraphics[width=\linewidth]{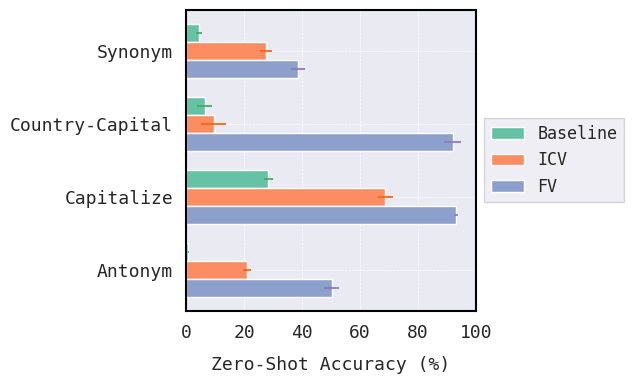}
        \caption{Functional tasks.}
        \label{fig:icv_vs_fv.functional}
    \end{subfigure}
    \begin{subfigure}[b]{0.49\textwidth}
        \centering
        \includegraphics[width=\linewidth]{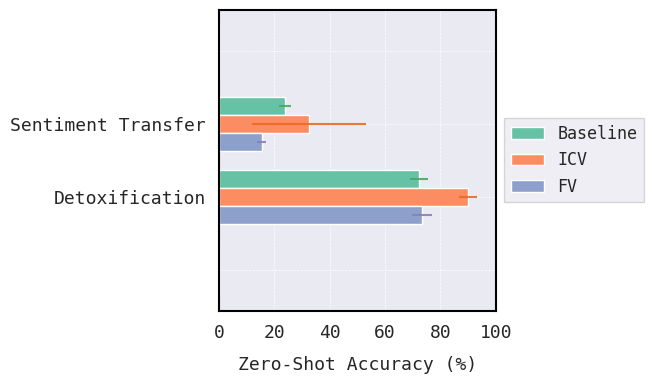}
        \caption{Behavioral tasks.}
        \label{fig:icv_vs_fv.behavioral}
    \end{subfigure}
    \caption{Comparative performance of steering methods applied to Llama 2-Chat (7B) across functional and behavioral tasks on the zero-shot setting, averaged across random seeds.  ICV results are based on the best-performing ICV for each task. Baseline performance corresponds to clean model performance on the same zero-shot prompt.}
    \label{fig:icv_vs_fv}
    \vspace{-3mm}
\end{figure}

\begin{figure}[H]
    \centering
    \begin{subfigure}{0.49\textwidth}
        \centering
        \includegraphics[width=\linewidth]{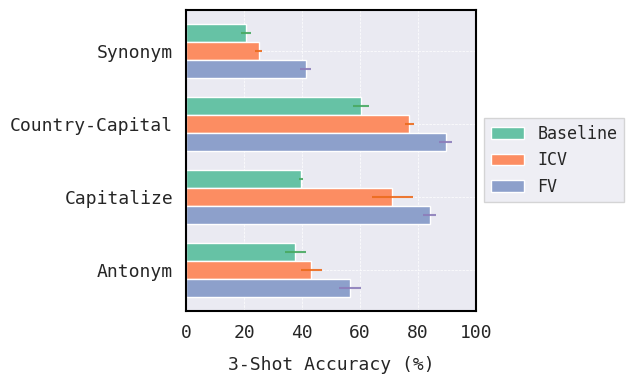}
        \caption{Shuffled-label 3-shot}
    \end{subfigure}
    \begin{subfigure}{0.49\textwidth}
        \centering
        \includegraphics[width=\linewidth]{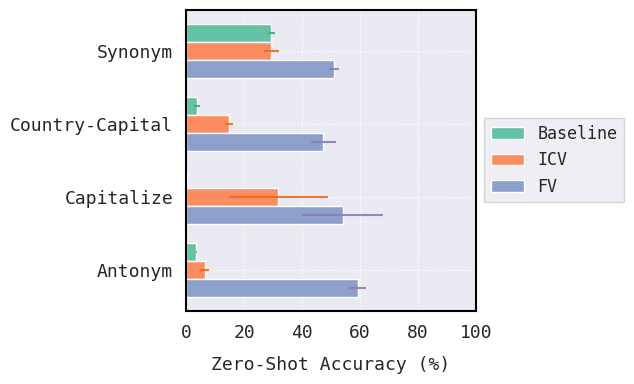}
        \caption{Natural text}
    \end{subfigure}
    \caption{Comparative performance of steering methods applied to Llama 2-Chat (7B) across functional tasks on various `OOD' settings, averaged across random seeds.  ICV results are based on the best-performing ICV for each task. Both vectors steer reasonably well in contexts similar to the in-context learning prompts from which they were extracted (see Appendix \ref{sec:experiment}), but both vectors, to varying degrees, have more difficulty generalizing to more out-of-distribution settings. Details about natural text prompting styles used can be found in Appendix \ref{sec:prompting_style}. }
    \label{fig:nshots}
    \vspace{-3mm}
\end{figure}

\textbf{Evaluation metrics.} We use several metrics to assess the effects of the two steering methods on the model. \emph{Accuracy} on functional tasks considers generations correct iff the first word generated matches the first word of the correct label. For natural text settings, it is sufficient for the ground truth label to be present anywhere in the generation. \emph{Behavioral shift classifiers} measure accuracy on behavioral tasks by the percentage of generations marked as demonstrating desired behavior (we use classifiers from prior literature -- see Appendix~\ref{sec:metrics}). To measure \emph{fluency}, we track three metrics used in prior works. Following \citet{meng2023locating}, we calculate generation entropy (GE), the weighted average of bi- and tri-gram entropies \citep{zhang2018generatinginformativediverseconversational}, as a measure of fluency. This score decreases as generations become more repetitive, a common failure mode of intervention steering methods \citep{meng2023locating}.

\section{Results}

This section discusses our steering performance results for the two families of tasks. Figure~\ref{fig:icv_vs_fv} summarizes our results.
We also discuss our investigations into why we observe these results. 

\subsection{Functional \& Behavioral Task Performance}
\label{sec:func_results}
\textbf{In-context vectors can steer for functional tasks, but function vectors outperform them.} Both steering methods can, generally, achieve substantial improvement over baseline accuracy across tasks.
Baseline accuracy corresponds to ICL performance given an `empty' prompt.
ICVs demonstrate surprisingly strong improvements in the 0-shot setting, though FVs performance exceeds ICVs performance by a significant margin on all functional tasks in all settings. FVs, therefore, appear to be particularly adept at steering precise behaviors in in-distribution contexts. 

\textbf{Function vectors struggle to capture high-level representations.} In both behavioral tasks, we observe better performance from ICVs than FVs at steering toward desired behavior. Function vectors, in fact, even steer \textit{away} from desired behavior on the sentiment transfer task. 
This indicates that function vectors are potentially capable of capturing high-level concepts conveyed in task-specific demonstrations, but do not do so well enough to reliably steer behavior in the desired direction.

However, the effects of ICVs can also be somewhat volatile; note the high variance in sentiment transfer performance in Figure~\ref{fig:icv_vs_fv.behavioral}.
ICVs are generally capable of steering effectively on behavioral tasks, but whether any steering capability emerges is highly dependent on the particular demonstration data used to extract the vector.
ICVs extracted from different task-specific demonstration data, but otherwise from the same number of demonstrations and applied to the model with the same vector strength, have vastly different effects on task performance.
This is evident from the sentiment transfer performance shown here; also see Appendix \ref{sec:task_demos} and Table \ref{tab:degenerated-fluency-table}. 

\subsection{Generalizability}
\label{sec:behav_results}

\begin{figure}[t]
    \begin{subfigure}[b]{0.39\textwidth}
    \includegraphics[width=0.99\linewidth]{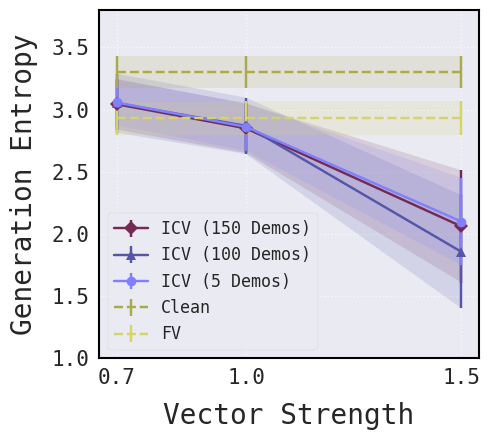}
    \caption{}
    \label{fig:fluency}
    \end{subfigure}
    \begin{subfigure}[b]{0.54\textwidth}
    \centering
    \includegraphics[width=0.99\linewidth]{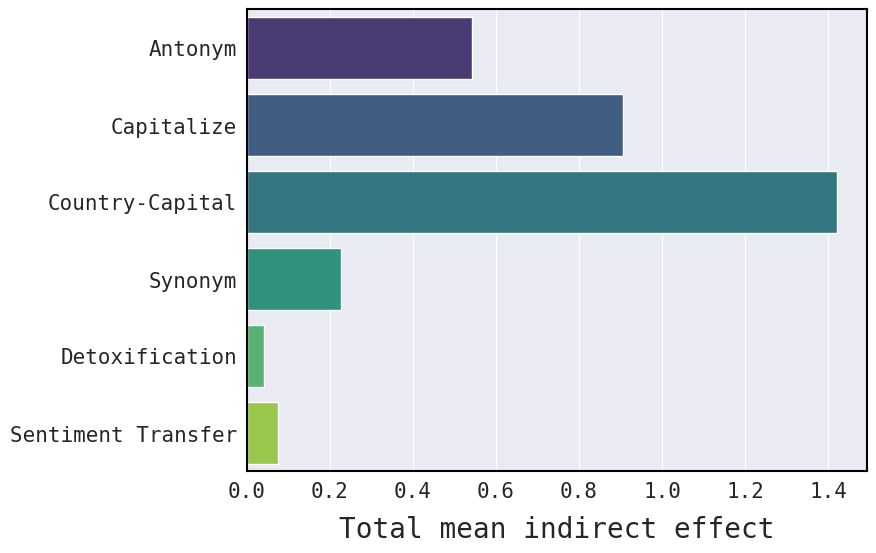}
    \caption{}
    \label{fig:cie.fvs}
    \end{subfigure}
    \caption{(a) Generation entropy (GE) scores by vector steering method, averaged across zero-shot tasks. 
    Vector strength is not varied for FVs. (b) The total mean CIE across the top 20 most implicated attention heads for each task. The total mean CIE represents the magnitude of impact the most influential attention heads have for a given task. Total mean CIE is strongly correlated with FV task performance.
    } 
\end{figure}
\textbf{ICVs have unpredictable effects across settings.} FVs and ICVs evaluated on $n$-shot settings are extracted using demonstration data that resembles the $n$-shot evaluation setting (illustrated in Appendix \ref{sec:prompting_style}). Transferring these same vectors to natural text settings, both vectors experience general drops in accuracy, but FVs still perform considerably better than the baseline on all tasks in the OOD natural text setting, indicating that FVs can transfer well to new settings. ICVs, meanwhile, lose steering abilities on some tasks, improving very little or not at all on baseline performance, while retaining steering abilities on others. 

\textbf{ICV-steered models produce less fluent generations than FV-steered models.} For both ICVs and FVs, the best performing vectors typically have only minor effects on overall model fluency, though ICVs are sensitive to vector strength used and are prone to causing degeneration in model outputs if the selected strength is too high. It is difficult to identify any particular ICV vector strength as ideal (producing strong steering effects without significant fluency degeneration) across tasks; higher vector strengths are often necessary to produce noticable improvements in task performance, and the maximum vector strength before significant coherence degradations occur varies widely between tasks as illustrated in Figure~\ref{fig:fluency}.
Note that the strength is not varied for FVs.

\subsection{FV Task Performance is Highly Correlated With Task-Specific CIE}
FVs are constructed by first identifying attention heads with high causal indirect effect (CIE) on model's performance towards accurately completing the desired task.
In other words, the construction of FVs relies on the assumption that there are a subset of attention heads whose activations collectively encode the in-context learning task.
We hypothesize that variability in FVs' performance on different tasks originates from how well the aforementioned assumption is satisfied for different tasks.
Indeed, in Figure~\ref{fig:cie.fvs}, it can be observed that FVs perform well on tasks with high mean CIE (e.g., country-capital), while they do poorly on tasks with low mean CIE (e.g., detoxification). This indicates that FV steering ability depends on the existence of a set of attention heads that have a significant effect on recovering task performance. The absence of such a set of heads for certain tasks suggests that execution of said tasks is not mediated by attention alone.

\begin{figure}
    \centering
    \begin{subfigure}[b]{0.48\textwidth}
    \centering
    \includegraphics[width=0.95\linewidth]{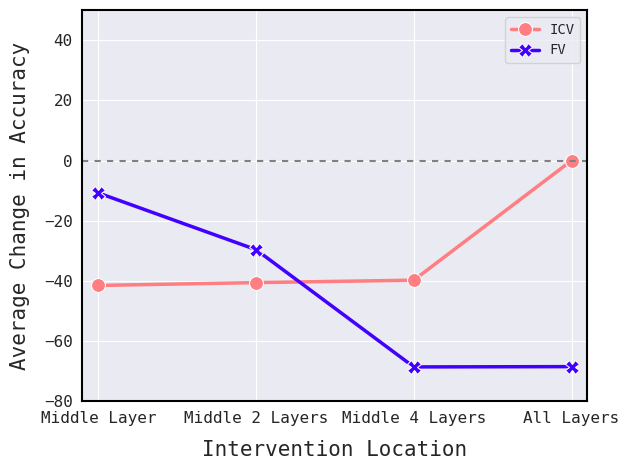}
    \caption{Functional tasks.}
    \label{fig:func-ablation}
    \end{subfigure}
    \begin{subfigure}[b]{0.48\textwidth}
    \centering
    \includegraphics[width=0.95\linewidth]{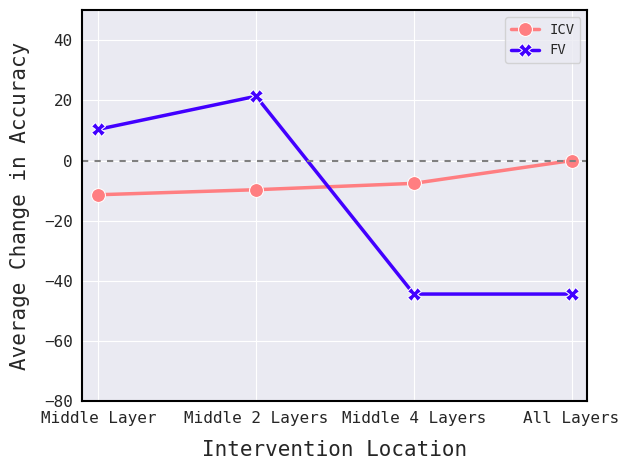}
    \caption{Behavioral tasks.}
    \label{fig:behav-ablation}  
    \end{subfigure} 
    \caption{Changes in overall task performance for FVs and ICVs by intervention location. We experiment with adding FVs to multiple layers, rather than a single layer, as well as adding ICVs at single layers rather than all layers. Note that all changes are relative to average task performance. On functional tasks, all changes in accuracy are negative, indicating no alternative intervention location improved on original FV or ICV task performance. However, for behavioral tasks, FV intervention accuracy at middle layers does show improvement.}
    \label{fig:ablation}
\end{figure}

\subsection{Ablation Study}
\begin{wrapfigure}{r}{0.4\textwidth}
    \centering
    \includegraphics[width=0.95\linewidth]{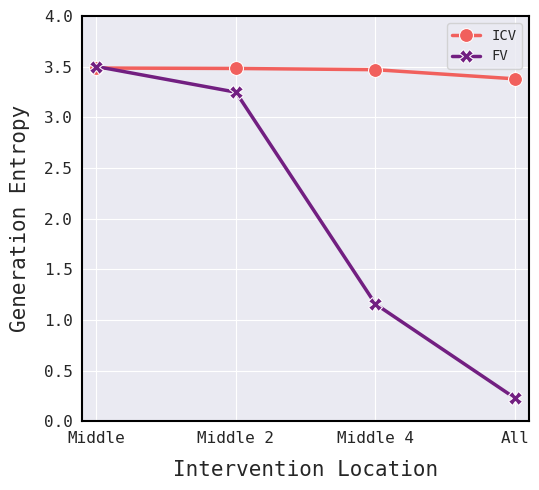}
    \caption{Generation entropy by intervention location, averaged across behavioral tasks, for FVs and ICVs.}
    \label{fig:flu-ablation}
    \vspace{-2em}
\end{wrapfigure}
We conduct an ablation study to determine the impact the location of intervention, as well as the breadth of intervention, has on the task performance we observe. Deviating from the original intervention procedures--recall that FVs are originally added to the layer $l \approx L/3$ and ICVs are added to all layers--we add FVs and ICVs at 1) the middle layer of the model, 2) the 2 middle layers of the model, and 3) the 4 middle layers of the model. We also experiment with adding FVs to all layers of the model.

As shown in Figure~\ref{fig:ablation}, ICVs universally steer less effectively when added to fewer layers of the model, consistent with the findings of \citet{liu2024incontext}. FV performance also declines significantly on functional tasks when added to multiple layers of the model. Model fluency also begins to collapse as FVs are added to more than 2 layers of the model, as seen in Figure~\ref{fig:flu-ablation}. However, varying intervention location and the number of layers intervened on can improve FV accuracy on behavioral tasks, as seen in Figure \ref{fig:behav-ablation}. As specified in \citet{todd2024function}, FVs are typically added at layer $L/3$, where $L$ is the number of layers in the model. Adding FVs to a later middle layer boosts average behavioral task performance by $10.36\%$, and adding the FV to 2 middle layers can improve behavioral task performance by up to $21.36\%$. This indicates that FVs do, indeed, capture some degree of behavioral task information that is useful for steering for desired behaviors. Additionally, this suggests that broader behavioral tasks may require more invasive steering interventions. That the location of these interventions differs meaningfully from the optimal location for functional task interventions may also be evidence that the mechanisms important for higher level concepts and behaviors in models exist in later middle layers.

\section{Discussion and Future Work}
Our results provide several insights into the characteristics of certain bottom-up and top-down methods for model steering. Our main takeaway is that \textbf{both studied methods claim to capture compact representations of in-context learning tasks that can be used in place of in-context learning, but we find that both methods are effective on only specific types of ICL tasks.} We also observe that, while both methods' steering effectiveness deflates in contexts different from the vector extraction setting, top-down methods may be particularly sensitive to shifts in setting and steer poorly as a result. Though the exact reasons for each method's respective limitations remain unclear, it is evident that neither method can presently serve as an effective substitute for in-context learning in general.

As for \textit{why} the methods studied are better at certain ICL tasks than others, we offer some hypotheses in this section. 
Our findings suggest that more surgical bottom-up methods are capable of capturing precise functions but cannot capture wider behaviors, likely because these behaviors are not mediated by only a few attention heads. The low overall CIE we observe for the top 20 attention heads implicated in behavioral tasks supports this hypothesis.

Similarly, top-down methods being unable to reliably perform fine-grained functional tasks can potentially be explained by the presumably small likelihood that the broad activation space from which they are extracted contains the fine mechanisms of a precise functional task within a single direction.
Generally speaking, the extent to which each kind of vector can steer effectively is constrained by whether the portion of latent space from which they are extracted actually contains a discoverable representation of the information needed to execute the desired behavior.
Determining whether certain representations exist within a specific portion of latent space is a central goal of interpretability. 
Developments in vector steering methods might lend researchers important clues to this end, but in the other direction, vector steering methods may not be viable as robust strategies for LLM control until the mechanisms within models that are necessary for certain behaviors are better understood. 

It is also clear that inconsistent evaluation setups across steering vector studies may prevent important limitations of steering methods from being discovered and thoroughly investigated. We thus stress the need for a unified benchmark for evaluating the performance and potential side effects of intervention steering methods that encompasses a wide range of highly diverse tasks and extends to several different settings to evaluate generalization behavior. Such a benchmark would allow for more thorough evaluation of intervention steering performance and better facilitate comparison between disparate methods.  

There are several obvious limitations of this work. 
We only compare FVs and ICVs on a relatively small set of functional and behavioral tasks, so it is possible our results may not generalize to an expanded set of tasks. 
Similarly, since our work only considers two steering methods, 
our findings ought to be treated cautiously and may not generalize to all ``bottom-up'' or ``top-down" approaches for vector steering or interpretability. 
In spite of these clear limitations, our work provides a signal for future investigations of bottom-up and top-down interpretability methods, as well as evidence of each approach's strengths and potential limitations.

Expanding the scope of this study to other tasks, other models and other steering methods is an obvious avenue for future work. 
Furthermore, this work makes it clear that existing theories of in-context learning in LLMs are incomplete.
Developing explanations for the variable performance of FVs and ICVs, as well as the behavior of bottom-up and top-down steering methods in general, may help improve our understanding of in-context learning in LLMs~\citep[][Section 2.1]{anwar2024foundationalchallengesassuringalignment}.

\begin{ack}
MB did this work as part of SPAR in summer 2024.
UA is supported by OpenPhil AI Fellowship and Vitalik Buterin Fellowship in AI Existential Safety.
\end{ack}

\bibliography{references}     
\bibliographystyle{abbrvnat}  

\clearpage
\appendix

\section{Related Works}

\subsection{Vector Steering}
\label{sec:steering_works}

Vector steering refers to a method of targeted intervention on language models' intermediate activations at inference time with computed "steering vectors" designed to elicit a desired behavior. Vector steering interventions generally do not require compute-intensive optimization as in fine-tuning, nor do they require the addition of tokens to the context to elicit desired behavior. Steering vectors are relatively inexpensive to compute, inexpensive to apply at inference time, and have been shown to be effective at steering model outputs on a variety of tasks. They thus represent an exciting frontier in AI safety and alignment research, as they can be deployed on top of fine-tuned models, such as Llama 2 \citep{touvron2023llama2openfoundation} to improve alignment with human values \citep{meng2023locating, wolf2020huggingface}. 

Many different methods of computing steering vectors exist \citep{panickssery2024steeringllama2contrastive, liu2024incontext, todd2024function, turner2024activationadditionsteeringlanguage, zou2023representation, hendel2023incontextlearningcreatestask, li2024incontextlearningstatevector} . Methods like Contrastive Activation Addition presented by \citet{panickssery2024steeringllama2contrastive} and In-Context Vectors presented by \citet{liu2024incontext} generate steering vectors by averaging or taking the principal direction of the differences in activations between pairs of positive and negative demonstrations of a particular behavior. Other methods such as Function Vectors \citep{todd2024function} and Task Vectors \citep{hendel2023incontextlearningcreatestask} perform more in-depth analysis of causally implicated components of the model, such as important attention heads, to extract steering vectors.

\subsection{Evaluations of Interpretability Techniques}

Interpretability methods for large language models are often claimed to be more reliable and generalizable than they actually are in practice \citep{anwar2024foundationalchallengesassuringalignment}. Thus, it is important for proposed interpretability techniques to be evaluated for robustness in diverse settings where they were not previously evaluated. Existing work investigating the reliability of steering vectors \citep{tan2024analyzinggeneralizationreliabilitysteering} find that steering vectors fail to generalize well on out-of-distribution settings in practice. However, to the best of our knowledge, no work has investigated the generalization behavior of steering vectors resulting from different vector extraction protocols, though understanding their relative strengths and weaknesses is crucial for designing more robust methods for model control and understanding.   

\section{Experimental Details}
\label{sec:experiment}
In this section, we provide details about the vector extraction process and evaluation procedure. 

\subsection{Prompting Styles}
\label{sec:prompting_style}
We prompt models at various stages of our procedure with zero-shot, few-shot, shuffled-label few-shot, and natural text prompting. Zero-shot prompts use the following template: \texttt{Q: $x_q$ \textbackslash nA:}, where $x_q$ is the query input. Similarly, for few-shot demonstrations, each input-output pair $(x, y)$ demonstrating the ICL task takes the form \texttt{Q: $x$\textbackslash n A: $y$\textbackslash n\textbackslash n}. The final few-shot prompt concatenates these $n$ demonstrations with the zero-shot form above. \citet{todd2024function} use shuffled-label few shot prompts as uninformative ICL prompts for function vector extraction and few-shot evaluation. To restate their formulation, for an $n$-shot shuffled-label prompt, $n$ ICL pairs $(x_i, y_i)$ are sampled from the task dataset; labels across all pairs are then shuffled such that each $x_i$ is paired with some random $\Tilde{y}_{i}$ in the set of labels $\{y_1, \ldots, y_n\}$. This removes the systematic relationship within input-output pairs while conditioning the model to respond in the form demonstrated by the prompt. These $n$ pairs are then concatenated with the query input $x_q$ for assessing the model to form a prompt, following the templates described above. Natural text prompt templates are sampled directly from \citep[][Appendix F]{todd2024function}. These templates situate the query input $x_q$ within a natural text sentence to elicit the desired completion.

\cite{todd2024function} uses shuffled-label few shot prompts as uninformative ICL prompts for function vector extraction and few-shot evaluation. Restating their formulation, for an $n$-shot shuffled-label prompt, $n$ ICL pairs $(x_i, y_i)$ are sampled from the task dataset; labels across all pairs are then shuffled such that each $x_i$ is paired with some random $\Tilde{y}_{i}$ in the set of labels $\{y_1, \ldots, y_n\}$. This removes the systematic relationship within input-output pairs while conditioning the model to respond in the form demonstrated by the prompt. These $n$ pairs are then concatenated with the query input $x_q$ for assessing the model to form a prompt. 

\subsection{In-Context Vector Extraction} 
\label{sec:icv_extract}
We provide a set of $k$ contrast pairs $X_{demos}=\{(x_i, y_i)|i=1, \ldots, k\}$ demonstrating target behavior; we then sweep across values of $k$ to find the optimal number of demonstrations for each task. Given a pair $(a, b)$ where $b$ is the desired transformation of $a$ corresponding to the given task, and a random uninformative 1- to 5-word string $c$, the resulting demonstration pair $(x_i, y_i)$ is
\begin{align}
    \label{icv:prompt}
    \text{(\texttt{Q:\{$a$\}\textbackslash nA:\{$c$\}, Q:\{$a$\}\textbackslash nA:\{$b$\}})}
\end{align}
We use this demonstration format on all \textit{functional tasks} to capture the QA setting in which we evaluate functional tasks as well as the desired mapping from $a$ to $b$ in a way similar to the demonstration format used for function vectors. $x_i$ is a negative example of the desired behavior, with no underlying mapping between $a$ and $c$, while $y_i$ demonstrates the desired output and underlying task mapping $a$ to $b$. We sample $a$ and $b$ from train splits of the task-specific dataset and $c$ from \cite{vogel2024repeng}. We assess several styles of demonstration and find this particular style of demonstration tends to work best in practice for extracting strong vectors on functional tasks. 

For behavioral tasks, we use the demonstration format $(x_i, y_i) = (a, b)$, closely following \citep{liu2024incontext}. We find this format is least disruptive of model fluency while still recovering steering performance on behavioral tasks. 

The resulting in-context vector is added to the hidden state residual stream at \textit{all} layers and every token position as described in \cite{liu2024incontext}.

\subsection{Function Vector Extraction} 
\label{sec:func_extract}
We closely follow the procedure for extracting function vectors as described in \citet{todd2024function}. To summarize this procedure: we first compute the task-conditioned mean activation of each attention head over 100 10-shot prompts in the form demonstrating the desired ICL task, then identifying a set of $k$ attention heads with the greatest average indirect effect towards recovering the correct answer given 25 shuffled-label (see above) 10 shot prompts. The number of heads to use in this procedure varies by model but scales approximately proportional to the number of attention heads in the model; for Llama 2 (7B), on which we run the majority of our experiments, we follow \citet{todd2024function} in using $k=20$ heads. Pythia (6.9B) \citep{biderman2023pythiasuiteanalyzinglarge} contains the same number of attention heads as Llama 2 (7B) and as such we use the same $k$ for its function vectors. 

Function vectors are added to a single layer $l$ at influence time, roughly $L/3$ where $L$ is the number of layers in the model, following \citet{todd2024function}. For Llama 2 (7B) and Pythia (6.9B), we set $l=11$. 

\subsection{Evaluation Procedure} 
\label{sec:eval_details}
For evaluation on functional tasks, we use the zero-shot, 3-shot, and natural text prompting styles discussed above. For evaluation on behavioral tasks, we use the zero-shot prompting style discussed above. Note that we depart from \citet{liu2024incontext} in our evaluation of detoxification and sentiment transfer tasks here. \citet{liu2024incontext} prompt by appending \texttt{ Paraphrase:} to the query inputs to encourage models to reword the input sentence with the desired shift in style or behavior. We instead elect to continue using the same prompting style for these behavioral tasks as functional tasks, primarily because we wish to observe steering behaviors without any additional influence from prompting which might interfere with vector steering effects.

We also sweep across vector strengths and numbers of demonstration samples for ICVs to determine the best performing ICV for each task. We do not perform the same sweeps for FVs. \citet{todd2024function} do not claim that varying FV strength is possible, and our attempts to do so resulted in incoherent generations. Preliminary sweeps over the size of shuffled-label prompt dataset used to determine average indirect effect required prohibitively large additional compute resources and did not result in any significant effects on performance that warranted further experimentation.

\subsection{Evaluation Metric Details}
\label{sec:metrics}
We use several evaluation metrics to assess both the positive and negative effects of the two steering methods on the model. 

\textbf{Accuracy:} We calculate accuracy of each model generation on functional tasks by \textit{first word score}, where generations are marked correct iff the first word of the generation matches the first word of the correct label. We report this as our primary accuracy metric. For natural text settings, where the nature of some prompts encourages articles to be generated first, we mark the generation correct if the ground truth label is present anywhere in the generation.

\textbf{Fluency:} We calculate the weighted average of bi- and tri-gram entropies \citep{zhang2018generatinginformativediverseconversational}, which we refer to as generation entropy (GE), following \citet{meng2023locating}. This score decreases as generations become more repetitive, a common failure mode of intervention steering methods \citep{meng2023locating}. We also evaluate sentence diversity using Dist-1 and Dist-2 metrics \citep{li-etal-2016-diversity}, which measure diversity and fluency by counting unique uni- and bi-grams in a sentence. 

\textbf{Behavioral Shift:} We measure success on behavioral tasks by the percentage of generations that are classified as demonstrating desired behavior. For the detoxification task, following \citet{liu2024incontext}, we use ParlAI's safety classifier \citep{miller2017parlai, xu2021recipessafetyopendomainchatbots} to evaluate generation safety on the detoxification task. We mark a response unsafe if the classifier labels it unsafe with probability greater than $0.9$; the behavioral shift score is the percentage of generations that were not marked unsafe. For sentiment transfer, following \citet{turner2024activationadditionsteeringlanguage}, we use SiEBERT \citep{hartmann2023} to classify sentiment for the sentiment transfer task. Generations are marked correct if they are classified as having positive sentiment.    
 
\section{Datasets}
\label{sec:datasets}
All datasets for functional tasks (antonym, capitalization, country-capital, and synonym) are sourced from \citep{todd2024function}. We provide details about the datasets used for behavioral tasks below.

\textbf{Detoxification.} We construct our detoxification dataset by first deriving the most toxic examples from from the ParaDetox dataset \citep{logacheva-etal-2022-paradetox}, as graded by GPT-4 \citep{openai2024gpt4technicalreport}. We then construct 1057 contrast pairs by prompting GPT-4 to rewrite each toxic sentence such that all insulting, offensive, and discriminatory content is removed, changing the original messaging to be more respectful if necessary. Toxic and detoxified sentences are paired to form a dataset of input-output pairs demonstrating the detoxification task.

\textbf{Sentiment Transfer.} We construct a dataset of 1000 negative sentiment to positive sentiment contrast pairs from the Yelp reviews full star dataset \citep{zhang2016characterlevelconvolutionalnetworkstext}. We select 1000 1-star reviews from the Yelp dataset, then prompt GPT-4 to first rewrite the review to fit within a 25-word limit and then rewrite the review to convey positive sentiment. The negative-positive sentiment rewrites are then paired.

\section{ICV Task Demonstration Forms}
\label{sec:task_demos}

In this section, we discuss our decisions to use the ICV demonstration formats described in Appendix \ref{sec:experiment} in more detail. We experiment with several templates for ICV task demonstrations on both categories of task and find that ICVs show significant sensitivity to the style of demonstration used for vector extraction. 

Initial experiments with ICVs on functional tasks were performed adhering closely to the ICV extraction procedure described in \citep{liu2024incontext}, where 5 contrast pairs $(x_i, y_i)$, each consisting of a negative example $x_i$ and positive example $y_i$ of the desired behavior, are used as demonstrations to extract the vector. Curiously, ICV performance on the capitalization task (shown in Table \ref{tab:demo-form-comparison} as "Style 1" for this demonstration form) showed no significant improvement on baseline accuracy. \citet{liu2024incontext} demonstrate that ICVs are capable of steering text formatting, and first letter capitalization in particular, motivating additional experimentation with demonstration style to investigate whether performance can be recovered with a more well-formed demonstration format---particularly, more precise "postive" and "negative" demonstrations. Indeed, adjusting the demonstration form to that shown in Figure \ref{icv:prompt}, referred to in Table \ref{tab:demo-form-comparison} as "Style 2", recovered significant ICV steering ability on zero-shot and 3-shot settings without major fluency side-effects. 

\begin{table}[h]
\centering
\small
\begin{tabular}{l rrrr}
\toprule
\textbf{Metric} & Base & FV & ICV (Style 1) & ICV (Style 2) \\
\midrule
Accuracy & 28.8\% & \textbf{97.1\%} & 25.9\% & 66.5\% \\
Dist-1 & 87.2 & \textbf{87.9} & 87.8 & 85.8 \\
Dist-2 & 87.4 & \textbf{89.7} & 87.2 & 86.8 \\
GE & 3.226 & \textbf{3.255} & 3.229 & 3.213 \\
\bottomrule
\end{tabular}
\caption{Performance of each steering method, including ICVs extracted using 2 different styles of demonstrations, on the capitalization task. Results are from Llama-2-7b-chat. Style 2 shows a marked improvement over Style 1, which performs on par with baseline model performance.}
\label{tab:demo-form-comparison}
\end{table}

\begin{table}[h]
\small
\centering
\setlength{\extrarowheight}{2pt}
\caption{Example outputs generated with ICV-steered Pythia (6.9B) on detoxification and ICV-steered Llama 2-Chat (7B) on sentiment transfer tasks, using demonstrations matching \ref{icv:prompt}. Generations are shown in \textcolor{darkgreen}{green}.}
\begin{tabular}{p{0.95\textwidth}}
\hline
\textbf{Q:} until then , as i said, they can kiss my ass . \\
\textbf{A:} \textcolor{darkgreen}{\textbackslash n\textbackslash nA:\textbackslash n\textbackslash nA:\textbackslash n\textbackslash nA:\textbackslash n\textbackslash nA:} \\
\hline
\textbf{Q:} only a total crook would do what these scumbag libs are planning . \\
\textbf{A:} \textcolor{darkgreen}{\textbackslash n\textbackslash n.\textbackslash n\textbackslash nHowever, the law.\textbackslash n\textbackslash nHowever, the law}\\
\hline
\textbf{Q:} Called for a 3-day car rental; they refused to price match, quoted higher rates than online. Clueless staff, terrible customer service, avoid this place!  \\
\textbf{A:} \textcolor{darkgreen}{great deals, friendly and knowledgeable staff, highly recommended!\textbackslash n\textbackslash n\textbackslash n} \\
\hline
\textbf{Q:} Horrible service and inconsistent food ruined my friend's birthday dinner. Bland spring rolls, wrong and overly spicy Moak Pa, and a rude waitress. Never returning. \\
\textbf{A:} \textcolor{darkgreen}{flavors.\textbackslash n\textbackslash n\textbackslash n\textbackslash n\textbackslash n\textbackslash n\textbackslash n\textbackslash n\textbackslash n\textbackslash n\textbackslash n\textbackslash n\textbackslash n\textbackslash n} \\
\hline
\end{tabular}
\label{tab:icv-example-outputs}
\end{table}

\begin{table}
\small
\caption{Performance of each steering method applied to Pythia (6.9B) and Llama 2-Chat (7B) on detoxification and sentiment transfer tasks. ICV results are taken from the best performing ICV on average, extracted with demonstrations in the form shown in Figure \ref{icv:prompt}. The worst fluency scores are listed in red.}
\begin{center}
\begin{tabularx}{\textwidth}{lXXXXXX}
    \toprule
    & \multicolumn{3}{c}{Detoxification (Llama-2-7b)} & \multicolumn{3}{c}{Sentiment Transfer (Llama-2-7b-chat)} \\
    \cmidrule(lr){2-4} \cmidrule(lr){5-7}
    \textbf{Metric} & Base & FV & ICV & Base & FV & ICV \\
    \midrule
    Beh. Shift (\%) $\uparrow$& 67.19 $\pm$ 0.0& $40.5 \pm 33.1$ & $\mathbf{96.2} \pm 0.7$ & $23.81 \pm 2.1$ & $15.4 \pm 1.6$ & $\mathbf{86.5} \pm 4.1$ \\
    Dist-1 $\uparrow$ & $\mathbf{83.2} \pm 0.4$ & $82.1 \pm 1.2$ & \textcolor{red}{$47.4 \pm 3.9$} & $\mathbf{97.2} \pm 0.1$ & $96.9 \pm 0.4$ & \textcolor{red}{$75.8 \pm 7.6$} \\
    Dist-2 $\uparrow$ & $\mathbf{84.5} \pm 0.2$ & $79.4 \pm 2.8$ & \textcolor{red}{$50.6 \pm 4.4$} & $\mathbf{91.7} \pm 0.5$ & $91.6 \pm 0.5$ & \textcolor{red}{$71.8 \pm 7.9$} \\
    GE $\uparrow$ & $\mathbf{3.49} \pm 0.00$ & $3.06 \pm 0.57$ & \textcolor{red}{$2.74 \pm 0.12$} & $\mathbf{3.52} \pm 0.04$ & $3.45 \pm 0.06$ & \textcolor{red}{$2.37 \pm 0.34$} \\
    \bottomrule
\end{tabularx}
\end{center}
\label{tab:degenerated-fluency-table}
\end{table}

On the other hand, using the demonstration form shown in \ref{icv:prompt} for vector steering on behavioral tasks can cause total breakdown in model language capabilities with even small vector strengths (though these vectors do still successfully encode task information, to the extent that severely disfluent generations can still be evaluated). Examples of generations on behavioral tasks are shown in Table \ref{tab:icv-example-outputs}, and evaluations of behavioral task performance with this demonstration form can be seen in Table \ref{tab:degenerated-fluency-table}. Generations are only graded for accuracy if their generation entropy scores are above $2.0$ (by this criteria, only the third row in Table \ref{tab:icv-example-outputs} would be scored for accuracy; the rest would be ignored), and evaluation statistics are only considered for ICVs where at least 60\% of generations could be graded. Returning to the demonstration style used in \citep{liu2024incontext} improved fluency significantly, though steering performance was somewhat reduced; these results are reported in Section \ref{sec:behav_results}.

Overall, ICVs appear to overfit to the style of demonstration used for vector extraction, working well in settings that very closely resemble the form of demonstrations used for extraction, but losing steering abilities and becoming prone to causing language degeneration in more out-of-distribution settings.

\end{document}